\pdfoutput=1

\documentclass[11pt]{article}

\usepackage{acl}

\usepackage{times}
\usepackage{latexsym}
\usepackage{graphicx}

\usepackage{adjustbox}
\usepackage{booktabs}
\usepackage{arydshln}
\usepackage{multirow}
\usepackage{dirtytalk}
\usepackage{todonotes}
\usepackage{color}
\usepackage{caption}
\usepackage{amsmath, amssymb}

\DeclareCaptionType{exam}[Example][List of Examples]

\usepackage[T1]{fontenc}

\usepackage[utf8]{inputenc}

\usepackage{microtype}

\makeatletter
\newcommand\footnoteref[1]{\protected@xdef\@thefnmark{\ref{#1}}\@footnotemark}
\makeatother

\makeatletter
\def\adl@drawiv#1#2#3{%
        \hskip.5\tabcolsep
        \xleaders#3{#2.5\@tempdimb #1{1}#2.5\@tempdimb}%
                #2\z@ plus1fil minus1fil\relax
        \hskip.5\tabcolsep}
\newcommand{\cdashlinelr}[1]{%
  \noalign{\vskip\aboverulesep
           \global\let\@dashdrawstore\adl@draw
           \global\let\adl@draw\adl@drawiv}
  \cdashline{#1}
  \noalign{\global\let\adl@draw\@dashdrawstore
           \vskip\belowrulesep}}
\makeatother

\title{End-to-end Multilingual Coreference Resolution with Mention Head Prediction}

\author{Ondřej Pražák \and Miloslav Konopík\\[0.5em]
\tt{\{ondfa,konopik\}@kiv.zcu.cz}\\[0.5em]
Department of Computer Science and Engineering, \\
NTIS -- New Technologies for the Information Society, \\
Faculty of Applied Sciences, University of West Bohemia, Technick\'a 8, 306 14 Plze\v{n} \\
Czech Republic \\
}

\begin{document}
\maketitle

\begin{abstract}
This paper describes our approach to the CRAC 2022 Shared Task on Multilingual Coreference Resolution. Our model is based on a state-of-the-art end-to-end coreference resolution system. Apart from joined multilingual training, we improved our results with mention head prediction. We also tried to integrate dependency information into our model. Our system ended up in $3^{rd}$ place. Moreover, we reached the best performance on two datasets out of 13.

\end{abstract}

\section{Introduction}
Coreference resolution is the task of finding language expressions that refer to the same real-world entity (antecedent) of a given text. Sometimes the corefering expressions can come from a single sentence. However, the expressions can be one or more sentences apart as well. It is necessary to see the whole document in some hard cases to judge whether two expressions are corefering adequately. This task can be divided into two subtasks. Identifying entity mentions, and grouping the mentions together according to the real-world entity they refer to. The task of coreference resolution is closely related to anaphora resolution -- see \cite{Sukthanker2020} to compare these two tasks. 

This paper describes our approach to the CRAC 2022 Shared Task on Multilingual Coreference Resolution. The task is based on the CorefUD dataset \cite{corefud2022lrec}. The CorefUD corpus contains 13 different datasets for ten languages in a harmonized scheme. As the CorefUD is meant to be the extension of Universal Dependencies with coreference annotation, all the datasets in CorefUD are treebanks. For some languages, human annotators provided the dependency annotations. For others, the annotation is created automatically with a parser. The coreference annotation is built upon the dependencies. This means that the mentions are subtrees in the dependency tree and can be represented with the head. In fact, in some of the datasets, there are non-treelet mentions -- the mentions which do not form a single subtree. But even for these non-treelet mentions, a single headword is selected. There are some notable differences between the datasets. One of the most prominent ones is the presence of singletons. Singletons are clusters that contain only one mention. Singletons are not present in any coreference relation. However, they are annotated as mentions. For details about the dataset, please see \citet{corefud2022lrec} or \citet{Nedoluzhko2021}. The task was simplified to predict only non-singleton mentions and group them into entity clusters. 

For evaluation, the CorefUD scorer\footnote{\url{https://github.com/ufal/corefud-scorer}} is provided. The primary evaluation score is the CoNLL $F_1$ score with partial mention matching and singletons excluded. In the CorefUD scorer, a system mention matches a gold mention if all its words are included in the gold mention, and one of them is the key head. This means that the minimal correct span is the head, and it might be beneficial to predict mentions as only the heads.

\begin{table*}[]
    \centering
    \begin{tabular}{lcccccc}
    \toprule
        \multirow{2}{*}{CorefUD dataset} & \multicolumn{6}{c}{Total size} \\
        \cline{2-7}
         & docs & sents & words & empty & singletons & discont. \\
         \midrule
         Catalan-AnCora & 1550 & 16,678 & 488,379 &6,377 & 74.6\% & 0\% \\
         Czech-PDT & 3165 & 49,428 & 834,721 & 33,086 & 35.3\% & 3.1\% \\
         Czech-PCEDT & 2312 & 49,208 & 1,155,755 & 45,158 & 1.4\% & 4.1\% \\
         English-GUM & 150 & 7,408 & 134,474 & 0 & 75\% & 0\% \\
         English-ParCorFull & 19 & 543 & 10,798 & 0 & 6.1\% & 0.7\% \\
         French-Democrat & 126 & 13,054 & 284,823 & 0 & 81.8\% & 0\% \\
         German-ParCorFull & 19 & 543 & 10,602 & 0 & 5.8\% & 0.3\% \\
         German-PotsdamCC & 176 & 2,238 & 33,222 & 0 & 76.5\% & 6.3\% \\
         Hungarian-SzegedKoref & 400 & 8,820 & 123,976 & 4,849 & 7.9\% & 0.4\% \\
         Lithuanian-LCC & 100 & 1,714 & 37,014 & 0 & 11.2\% & 0\% \\
         Polish-PCC & 1828 & 35,874 & 538,891 & 864 & 82.6\% & 1.0\% \\
         Russian-RuCor & 181 & 9,035 & 156,636 & 0 & 2.5\% & 0.5\% \\
         Spanish-AnCora & 1635 & 17,662 & 517,258 & 8,111 & 73.4\% & 0\% \\
         \bottomrule
    \end{tabular}
    \caption{Dataset Statistics}
    \label{tab:stats}
\end{table*}

\section{Model}

Our model builds on the official transformer-based end-to-end baseline \cite{pravzak2021multilingual}. The underlying neural end-to-end coreference resolution model was originally proposed by \citet{lee-etal-2017-end}. The model predicts the antecedents directly from all possible mention spans without a previous discrete decision about mentions. In the training phase, it maximizes the marginal log-likelihood of all correct antecedents:

\begin{equation}
    J(D) = \log \prod_{i=1}^N \sum_{\hat{y} \in Y(i) \cap \texttt{GOLD}(i)}P(\hat{y})
\end{equation}\label{eq:loss}

where GOLD($i$) is the set of spans in the training data that are antecedents. 

The model achieves state-of-the-art performance on the OntoNotes dataset where singletons are not annotated. We believe the model is optimal for the CorefUD dataset as well since some of the datasets of the CorefUD do not contain singletons. Moreover, the evaluation metric ignores singletons, so it does not matter that the model is not able to predict them.

\paragraph{Employed Models}
We based our model on XLM Roberta large \cite{conneau2020unsupervised}, which is significantly larger than multilingual BERT \cite{bertpaper} used by the baseline. The number of parameters is provided in Table \ref{tab:num_params}. We also tried to use the best monolingual model for each language.

\begin{table}[]
    \centering
    \begin{tabular}{lcc}
    \toprule
         Model & Pretrained params & New params  \\
         \midrule
         mBERT & 180M & 40M \\
         XLM-R & 350M & 50M \\
         \bottomrule
    \end{tabular}
    \caption{Number of trainable parameters of the models}
    \label{tab:num_params}
\end{table}

\paragraph{Joined Model Pretraining}

As you can see from Table \ref{tab:num_params}, there are approximately 50 million parameters trained from scratch for XLM-R. For smaller datasets, it is practically impossible to train so many random parameters. To solve this issue, we first pre-train the model on the joined dataset and then fine-tune the model for a specific language. 

\paragraph{Heads Prediction}

As mentioned above, the official scorer uses min-span evaluation with head words as min spans. Because we do not know the rules used to select single mention head in the dataset, we decided to train to model to predict the heads instead of the whole spans to optimize the evaluation metric. Having all the useful information (even dependency trees), the model should learn the original rules for selecting the head.

The simplest way to  predict the mention heads would be to simply represent mention with its head word on the input. But this is not an ideal solution since multiple mentions can have the same head. If we represented a mention with only the head, some mentions would be joined, and their clusters would be merged.

To avoid this, we represent mention with the whole span, and just at the top of our model, we predict the head of each mention and output only the head word(s). This way, the mentions are represented with their spans when we build the clusters, and the clusters of two different mentions with the same head are not merged as in the case of the simple approach mentioned above.

We implemented two versions of the head prediction model. Both are implemented as separate classification heads on the top of our coreference resolution model.

The first model predicts the relative position of the head word(s) inside a span using the hidden representation of the span from the CR model. Output probabilities of head positions are obtained using sigmoid activation so the model can predict multiple heads even though there is only single head word in the gold data. This is particular optimization of the evaluation metric: If there are more words likely to be a head word of the span, it is statistically better to output all of them. 

The second model uses a binary classification of each span and head candidate pair, so again, there can be more head words of a single span predicted.

\paragraph{Trees}

We believe dependency information can help the model significantly, especially when manually annotated dependencies are provided (Czech PDT, for example). Moreover, the dependency information is necessary to find mention head.  

To encode syntactic information, we add to each token representation its path to \textit{ROOT} in the dependency tree. In more detail, we first set the maximum tree depth parameter and then concatenate Bert representations of all parents up to max depth with the embedding of the corresponding dependency relation. Thus the resulting tree structure representation has the size of $max\_tree\_depth \times (bert\_emb\_size + deprel\_emb\_size)$. This representation is then concatenated with bert embedding of each token.

\section{Training}

We trained all the models on NVIDIA A40 graphic cards using online learning (batch size 1 document). We limit the maximum sequence length to 6 non-overlapping segments of 512 tokens. During training, if the document is longer than $6 \times 512$ tokens, a random segment offset is sampled to take random continuous block of 6 segments, and the rest of them is discarded. During prediction, longer documents are split into independent sub-documents (for simplicity, non-overlapping again). We train a model for each dataset for approximately 80k updates in our monolingual experiments. For joined-pre-trained models, we use 80k steps for model pre-training on all the datasets and approximately 30k for fine-tunning on each dataset. Each training took from 8 to 20 hours.

\section{Results \& Discussion}

Results of several variants of our model are presented in Table \ref{tab:results}. 

\textit{Monoling} column shows the result of the monolingual model specific for each language. \textit{XLM-R} column presents results of XLM Roberta large trained for each dataset separately. \textit{Joined} is the joined model described in the Model section. The columns marked with \textit{+} mean the best model from all to the left, with the additional feature. \textit{+dev} means that the dev data part was added to training data, \textit{+S2H} is the model with mention head prediction described earlier. Both methods for mention head prediction have statistically equal performance (we cannot tell which one is better). The reported numbers are for the first one. The results in column \textit{+tree} correspond to adding the dependency structure as described.

\begin{table*}[]
    \centering
    \begin{adjustbox}{width=\linewidth,center}
    
    \begin{tabular}{@{}lllrrrrrrr@{}}
\toprule
Dataset/Model                       & \multicolumn{1}{l}{monolingual model name}  & \multicolumn{1}{l}{reference}                                      & \multicolumn{1}{c}{BASELINE} & \multicolumn{1}{c}{Monoling} & \multicolumn{1}{l}{XLM-R} & \multicolumn{1}{l}{joined} & \multicolumn{1}{l}{+dev} & \multicolumn{1}{l}{+S2H} & \multicolumn{1}{l}{+Tree} \\ \midrule
\multicolumn{1}{l|}{ca\_ancora}      & PlanTL-GOB-ES/roberta-base-ca   & \multicolumn{1}{r|}{\cite{armengol-estape-etal-2021-multilingual}} & 63.74                        & 69.61                        & 66.19                     & 68.81                      & \textbf{70.55}          & 69.91                   & 68.32                    \\
\multicolumn{1}{l|}{cs\_pcedt}       & Czert-B-base-cased              & \multicolumn{1}{r|}{\cite{sido2021czert}}                          & 70                           & 73.74                        & 73.55                     & 73.85                      & \textbf{74.07}          & 71.12                   & 73.61                    \\
\multicolumn{1}{l|}{cs\_pdt}         & Czert-B-base-cased              & \multicolumn{1}{r|}{\cite{sido2021czert}}                          & 67.27                        & 69.81                        & 70.99                     & 70.63                      & 71.49                   & \textbf{72.42}          & 70.99                    \\
\multicolumn{1}{l|}{de\_parcorfull}  & deepset/gbert-base              & \multicolumn{1}{r|}{\cite{chan2020german}}                         & 33.75                        & 43.04                        & 33.75                     & 68.91                      & \textbf{73.9}           & 68.3                    & 65.29                    \\
\multicolumn{1}{l|}{de\_potsdamcc}   & deepset/gbert-large             & \multicolumn{1}{r|}{\cite{chan2020german}}                         & 55.44                        & 58.81                        & 59.03                     & \textbf{70.35}             & 66.02                   & 68.68                   & 67.35                    \\
\multicolumn{1}{l|}{en\_gum}         & roberta-large                   & \multicolumn{1}{r|}{\cite{zhuang2021robustly}}                     & 62.59                        & 68                           & 66.27                     & 68.16                      & \textbf{68.31}          & 66.88                   & 67.39                    \\
\multicolumn{1}{l|}{en\_parcorfull}  & roberta-large                   & \multicolumn{1}{r|}{\cite{zhuang2021robustly}}                     & 36.44                        & 25.84                        & \textbf{36.44}            & 30.21                      & 31.9                    & 23.45                   & 40.05                    \\
\multicolumn{1}{l|}{es\_ancora}      & PlanTL-GOB-ES/roberta-large-bne & \multicolumn{1}{r|}{\cite{gutierrezfandino2022}}                   & 65.98                        & 60.12                        & 67.99                     & 71.24                      & 71.48                   & \textbf{72.32}          & 72.04                    \\
\multicolumn{1}{l|}{fr\_democrat}    & camembert/camembert-large       & \multicolumn{1}{r|}{\cite{martin2020camembert}}                    & 55.55                        & 56.76                        & 55.94                     & 59.8                       & 60.12                   & \textbf{61.39}          & 60.03                    \\
\multicolumn{1}{l|}{hu\_szegedkoref} & SZTAKI-HLT/hubert-base-cc       & \multicolumn{1}{r|}{\cite{nemeskey2021a}}                          & 52.35                        & 59.76                        & 60.68                     & 63.24                      & \textbf{65.01}          & 64.67                   & 62.77                    \\
\multicolumn{1}{l|}{lt\_lcc}         & EMBEDDIA/litlat-bert            & \multicolumn{1}{r|}{\cite{ulvcar2021training}}                     & 64.81                        & 66.93                        & 64.81                     & 66.34                      & \textbf{68.05}          & 67.49                   & 64.01                    \\
\multicolumn{1}{l|}{pl\_pcc}         & allegro/herbert-large-cased     & \multicolumn{1}{r|}{\cite{mroczkowskietal2021herbert}}             & 65.34                        & \textbf{75.2}                & 73.19                     & 73.66                      & 74.46                   & 74.56                   & 73.38                    \\
\multicolumn{1}{l|}{ru\_rucor}       & DeepPavlov/rubert-base-cased    & \multicolumn{1}{r|}{\cite{kuratov2019adaptation}}                  & 67.66                        & 69.33                        & \textbf{77.5}             & 75.5                       & 74.82                   & 76.02                   & 75.94                    \\ \hline
\multicolumn{1}{l|}{avg}                                 &                                  &      \multicolumn{1}{l|}{}                                                              & 58.53                        & 61.30                        & 62.03                     & 66.21                      & 66.94                   & 65.94                   & 65.94                    \\ \bottomrule
\end{tabular}
\end{adjustbox}
    \caption{Results}
    \label{tab:results}
\end{table*}

It is not surprising that employing a larger model (XLMR-R large or monolingual) significantly improved the performance of the baseline. The results of the joined model are much more interesting. We can see that for some smaller datasets (e. g. German), the performance gain is huge. But if we have a look at Table \ref{tab:num_params}, it makes sense because it is hard (or impossible) to train 50M parameters from scratch on a small dataset. It is also interesting that Polish is the only language where the monolingual model outperformed the joined model. But the reasons for this are probably straightforward. The polish dataset is one of the largest, so joined pre-training is not needed. Moreover there is a large monolingual model for Polish, so it is natural that it outperformed XLM-R large. For three datasets, there is a significant gain by employing mention head prediction. The difference should be even bigger when we add syntactic structure to the model. \footnote{The potential gain by outputting only mention heads can be found in \citet{crac2022findings}} Unfortunately, we did not manage to include this feature on time. From the results table, we can see that adding the trees does not help. In fact, it decreases performance significantly. We believe this is caused by some bug in our implementation, but we did not have enough time to correct it before the end of the competition.

\subsection{Comparison To Other Systems}

The comparison to other participating systems is shown in Table \ref{tab:comparison}. Our system ended up in $3^{rd}$ place. Surprisingly, although the winning system outperformed ours by a large margin on average, our system reached the best performance for two datasets (\emph{german\_parcor} and \emph{hungarian}). It would be interesting to look at both systems' differences to find out why.

\begin{table*}[]
    \centering
    \begin{adjustbox}{width=\linewidth,center}
    \begin{tabular}{lrlrrrrrrrrrrrrr}
\hline
\multicolumn{1}{c}{\#}      & \multicolumn{1}{c}{User}        & \multicolumn{1}{c}{avg} & \multicolumn{1}{c}{ca} & \multicolumn{1}{c}{cs\_pcedt} & \multicolumn{1}{c}{cs\_pdt} & \multicolumn{1}{c}{de\_pc} & \multicolumn{1}{c}{de\_pots} & \multicolumn{1}{c}{en\_gum} & \multicolumn{1}{c}{en\_pc} & \multicolumn{1}{c}{es} & \multicolumn{1}{c}{fr} & \multicolumn{1}{c}{hu} & \multicolumn{1}{c}{lt\_lcc} & \multicolumn{1}{c}{pl\_pcc} & \multicolumn{1}{c}{ru} \\ \hline
1                           & straka                          & \textbf{70.72}          & 78.18                  & \textbf{78.59}                & \textbf{77.69}              & 65.52                      & 70.69                        & 72.5                        & \textbf{39}                & \textbf{81.39}         & \textbf{65.27}         & 63.15                  & \textbf{69.92}              & 78.12                       & \textbf{79.34}         \\
2                           & straka-single-multil            & 69.56                   & \textbf{78.49}         & 78.49                         & 77.57                       & 59.94                      & \textbf{71.11}               & \textbf{73.2}               & 33.55                      & 80.8                   & 64.35                  & 63.38                  & 67.38                       & \textbf{78.32}              & 77.74                  \\
\textbf{3} & \textbf{ours} & 67.64                   & 70.55                  & 74.07                         & 72.42                       & \textbf{73.9}              & 68.68                        & 68.31                       & 31.9                       & 72.32                  & 61.39                  & \textbf{65.01}         & 68.05                       & 75.2                        & 77.5                   \\
4                           & straka-single-data              & 64.3                    & 76.34                  & 77.87                         & 76.76                       & 36.5                       & 56.65                        & 70.66                       & 23.48                      & 78.78                  & 64.94                  & 62.94                  & 61.32                       & 73.36                       & 76.26                  \\
5                           & berulasek                       & 59.72                   & 64.67                  & 70.56                         & 67.95                       & 38.5                       & 57.7                         & 63.07                       & 36.44                      & 66.61                  & 56.04                  & 55.02                  & 65.67                       & 65.99                       & 68.17                  \\
6                           & \textit{BASELINE}                           & 58.53                   & 63.74                  & 70                            & 67.27                       & 33.75                      & 55.44                        & 62.59                       & 36.44                      & 65.98                  & 55.55                  & 52.35                  & 64.81                       & 65.34                       & 67.66                  \\
7                           & Moravec                         & 55.05                   & 58.25                  & 68.19                         & 64.71                       & 31.86                      & 52.84                        & 59.15                       & 36.44                      & 62.01                  & 54.87                  & 52                     & 59.49                       & 63.4                        & 52.49                  \\
8                           & simple\_baseline                & 18.14                   & 15.58                  & 5.51                          & 9.48                        & 29.81                      & 19.41                        & 21.99                       & 11.37                      & 16.64                  & 21.74                  & 17                     & 27.53                       & 15.69                       & 24.06                  \\
9                           & k-sap                           & 5.9                     & 0                      & 0                             & 0                           & 0                          & 0                            & 0                           & 0                          & 0                      & 0                      & 0                      & 0                           & 76.67                       & 0                      \\ \hline
\end{tabular}
 \end{adjustbox}
    \caption{Comparison to Other Participating Systems}
    \label{tab:comparison}
\end{table*}

\section{Conclusion}

We successfully applied an end-to-end neural coreference resolution system to the CRAC 2022 shared task. There are two main outcomes of our work. \textbf{1)} Joined training helps a lot. Our experiments support the fulfillment of the goals of the CorefUD dataset to help the models by harmonizing the annotation schemas. \textbf{2)} For the official scoring metric, predicting only the mention heads increases performance. This means that syntactic structure helps to identify mentions. Of course, such evaluation is a bit artificial and does not reflect the real-world scenario, where we do not have the gold syntax. We also suggested injecting syntactic information into the model. Unfortunately, we did not manage to get any improvement with this approach. Our system ended up in $3^{rd}$ place. Moreover, we reached the best performance on two datasets out of 13.



\section*{Acknowledgements}
This work has been supported by Grant No. SGS-2022-016 Advanced methods of data processing and analysis. Computational resources were supplied by the project "e-Infrastruktura CZ" (e-INFRA CZ LM2018140) supported by the Ministry of Education, Youth and Sports of the Czech Republic.


\bibliography{anthology,custom}
\bibliographystyle{acl_natbib}


\end{document}